\PassOptionsToPackage{prologue,dvipsnames}{xcolor}
\documentclass{article}

\usepackage[preprint]{neurips_2023}

\usepackage[utf8]{inputenc} % allow utf-8 input
\usepackage[T1]{fontenc}    % use 8-bit T1 fonts
\usepackage{hyperref}       % hyperlinks
\usepackage{url}            % simple URL typesetting
\usepackage{booktabs}       % professional-quality tables
\usepackage{amsfonts}       % blackboard math symbols
\usepackage{nicefrac}       % compact symbols for 1/2, etc.
\usepackage{microtype}      % microtypography
\usepackage{xcolor}         % colors
\usepackage{enumitem}
\usepackage{graphicx}
\usepackage{tcolorbox}
\tcbuselibrary{breakable}
\usepackage{multirow}
\usepackage{makecell}
\usepackage{color}
\usepackage{enumitem}
\usepackage{amsmath}
\newenvironment{thmblock}[1][\textbf{Theorem}]{\begin{tcolorbox}[title=\textbf{#1}, colback=red!5,colframe=red!75!black, breakable]}{\end{tcolorbox}}

\newenvironment{propblock}[1][\textbf{Proposition}]{\begin{tcolorbox}
[title = \textbf{#1}, colback=Salmon!20, colframe=Salmon!90!Black, breakable]}{\end{tcolorbox}}

\newenvironment{colblock}[1][\textbf{Collary}]{\begin{tcolorbox}[colback=JungleGreen!10!Cerulean!15,colframe=CornflowerBlue!60!Black,title = \textbf{#1}, breakable]}{\end{tcolorbox}}
\title{Wider and Deeper LLM Networks \\ are Fairer LLM Evaluators}

\author{%
  Xinghua Zhang$^1$, Bowen Yu$^2$\thanks{\quad \small Correspondence to: Bowen Yu <yubowen.ybw@alibaba-inc.com>, Tingwen Liu <liutingwen@iie.ac.cn>.},  Haiyang Yu$^2$,\\ \textbf{Yangyu Lv$^2$, Tingwen Liu$^{1*}$, Fei Huang$^2$, Hongbo Xu$^1$, Yongbin Li$^2$} \\
$^1$ Institute of Information Engineering, Chinese Academy of Sciences\\ 
$^2$ Alibaba DAMO Academy\\  
\texttt{\{zhangxinghua,liutingwen,hbxu\}@iie.ac.cn},\\
\texttt{\{yubowen.ybw,yifei.yhy,yangyu.lyy,f.huang,shuide.lyb\}@alibaba-inc.com}\\
}

\begin{document}
\setcitestyle{numbers}

\maketitle

\begin{abstract}

Measuring the quality of responses generated by large language models (LLMs) is a challenging task, particularly when it comes to evaluating whether the response is aligned with human preference.
A novel approach involves using the LLM itself to make evaluation and stabilizing the results through multiple independent evaluations, similar to a single-layer narrow LLM network. This network consists of a fixed number of neurons, with each neuron being the same LLM.
In this paper, we draw upon the extensive research on deep neural networks to explore whether deeper and wider networks can lead to fairer evaluations. 
Specifically, inspired by the observation that different neurons in a neural network are responsible for detecting different concepts, we first adaptively generate as many neuron roles as possible for each evaluation sample. Each perspective corresponds to the role of a specific LLM neuron in the first layer.
In subsequent layers, we follow the idea that higher layers in deep networks are responsible for more comprehensive features, each layer receives representations from all neurons in the previous layer, integrating the locally learned evaluation information to obtain a more comprehensive evaluation result.
Interestingly, this network design resembles the process of academic paper reviewing, where each reviewer independently rates based on their preferences. Subsequently, through multiple discussions, they consider other reviewers' opinions to reach the final acceptance decision.
To validate the effectiveness of our method, we construct the largest and most diverse English evaluation benchmark LLMEval$^2$ for LLM evaluators, comprising 15 tasks, 8 abilities, and 2,553 samples. 
Experimental results demonstrate that a wider network (involving many reviewers) with 2 layers (one round of discussion) performs the best, improving kappa correlation coefficient from 0.28 to 0.34. 
% Evaluation results on real Chinese data demonstrate that our evaluator can save 80% of the annotation time for human evaluators, while achieving a 95% level of consistency with human assessments.
% The results of the actual testing demonstrate that our evaluator can save 80\% of the annotation time for human evaluators while achieving a performance level comparable to that of human annotators, reaching 95% accuracy.
We also leverage WideDeep to aid in the assessment of Chinese LLMs, which has accelerated the evaluation time by 4.6 times, resulting in a 60\% cost saving. WideDeep achieves a remarkable 93\% agreement level among humans\footnote{The data and code of this work is available at \url{https://github.com/AlibabaResearch/DAMO-ConvAI/tree/main/WideDeep}}.

\end{abstract}

\section{Introduction}

The rapid progress and remarkable achievements of large-scale pre-trained language models (LLMs) have catalyzed a revolutionary transformation in the realm of natural language processing~\citep{openai2022chatgpt,touvron2023llama,wei2023polylm}. 
These models have showcased substantial improvements across various applications, such as dialogue~\citep{zhao2023causal}, summarization~\citep{bhaskar2023prompted}, and code generation~\citep{chen2021evaluating}.
The majority of tasks involve open-ended, inherently subjective, and reference-free responses, rather than selecting from a fixed set of answers. 
Consequently, evaluating the correspondence of their generated responses with human intent becomes a challenge~\citep{saunders2022self}. 
Traditional automatic metrics such as BLEU~\citep{papineni-etal-2002-bleu} and ROUGE~\citep{lin-2004-rouge} have been shown to have relatively low correlation with human judgments, especially for open-ended generation tasks~\citep{liu2023gpteval},while human evaluation is often time-consuming and costly.
Thus, there is a growing demand for automated assessment methods that can consistently align with human judgments while being more efficient and cost-effective~\citep{jain2023self,li2023api,chang2023survey}.

\begin{figure}[t]
\centering
\includegraphics[width=1.0\columnwidth]{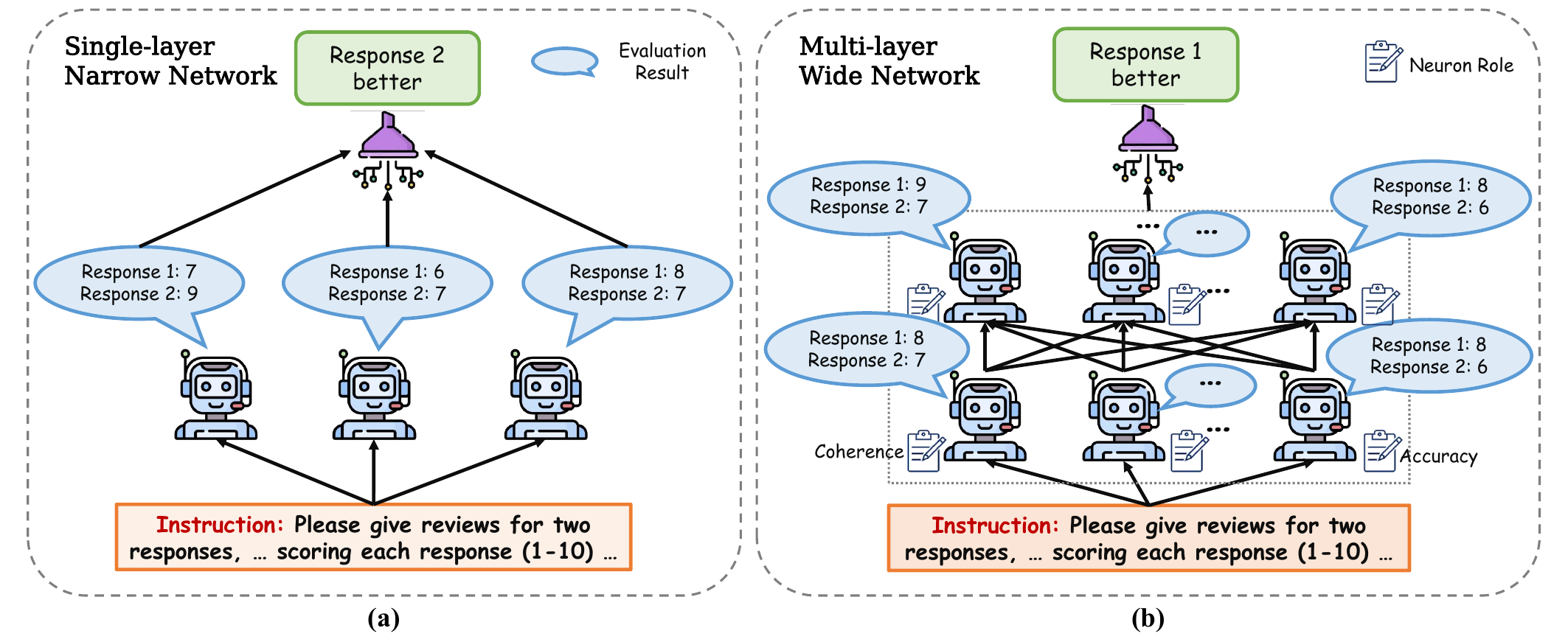} % Reduce the figure size so that it is slightly narrower than the column. Don't use precise values for figure width.This setup will avoid overfull boxes.
\caption{(a) Prior methods are single-layer LLM networks that combine assessments from a fixed number of LLM neurons. (b) In contrast, our method delves into the realm of wider and deeper multi-layer networks, where each neuron provides a distinct neuron role. }
\label{intro}
\end{figure}

Recent research has introduced the LLMs-as-evaluator paradigm, utilizing LLMs to compare candidate responses with the assumption that LLMs have learned to assign higher probabilities to high-quality and fluent texts~\citep{chen2023exploring, fu2023gptscore,ji2023exploring,kocmi2023large,wang2023chatgpt}.
% LLMs have revolutionized our understanding of the capabilities of machines, and achieved remarkable results in a wide range of NLP tasks. Moreover, recent studies~\citep{chen2023exploring, fu2023gptscore,ji2023exploring,kocmi2023large,wang2023chatgpt} have showcased the promising capabilities of LLMs in evaluating the quality of generated text. 
FairEval~\citep{wang2023large} finds that the ranking result of candidate responses can be easily altered by exchanging their order of appearance in prompt context. They swap the position of two responses for two rounds of scores and ensemble the results of multiple LLM runs in pursuit of result stability.
% FairEval also releases a benchmark containing 80 manually annotated samples.
Similarly, LLM-as-a-judge~\citep{zheng2023judging} also observes the position bias.
It swappes the order of two answers and retaines the evaluation score only if the results remain consistent in both orders. In cases of inconsistency after swapping, it declares a tie.
% Most of current studies conduct multiple independent evaluations to account for potential biases. 
Essentially, they regard each LLM as an individual neuron and construct single-layer narrow networks, aggregating evaluation scores from a limited quantity of LLMs. 
FairEval~\citep{wang2023large} identifies that the optimal performance is achieved when three LLM neurons are employed; an increase in the number of neurons leads to a decline in effectiveness.
Moreover, existing benchmarks for assessing LLMs' performance in evaluating text quality lack diverse evaluation capabilities.
For instance, the benchmark utilized by FairEval comprises only 80 samples. 
Thus, there is an urgent requirement for more comprehensive datasets that can holistically evaluate LLMs' ability to assess the quality of generated text.

% , i.e, FairEval使用的benchmark仅包含80 samples. There is a pressing need for more generalized datasets to comprehensively assess LLMs' evaluation of generated text.

In this paper, we first delve into the realm of deeper and wider LLM networks for LLM evaluation.
Systematic design has led to the development of deeper and wider neural networks, such as ResNets~\citep{he2016deep} for depth and ResNeXT~\citep{xie2017aggregated} for width. These advancements have resulted in enhanced learning and ultimately improved performance compared to relatively shallow and narrow networks~\citep{lee2017going}.
Therefore, we aim to increase the number of LLM neurons and layers that collaborate in the evaluation network, with the goal of creating a fairer LLM evaluator. 
It has been observed that different neurons in each layer of state-of-the-art deep networks match human-interpretable but distinct concepts~\citep{zhou2014object,zeiler2014visualizing,mahendran2015understanding,bau2018identifying,karpathy2015visualizing,radford2017learning,bau2020understanding}.
Moreover, the features in different layers focus on different views for samples~\citep{hao2019visualizing,qiao2019understanding,liu2021hit}.
For example, the features in lower layers tend to encode more local contents with basic syntactic representations in NLP.
Higher layers capture more complex semantics and usually produce higher-level semantic representations~\citep{gabeur2020multi,patrick2020support}.
However, in the evaluation network composed of different LLM neurons, we can only achieve forward computation and cannot update parameters as in deep neural networks where the different neurons are responsible for detecting different concepts and different layers abstract different granularity features through backpropagation. 
Therefore, in the network design, we artificially implement these two important characteristics.
Specifically, for each evaluation sample, we first ask the LLM about the candidate perspectives that could be used to assess the sample quality. Each perspective is explicitly injected into the evaluation process of each LLM neuron in the first layer as the concept that this neuron is responsible for detecting, outputting evaluation scores and reasons as the neuron's representation.
For subsequent layers in the multi-layer LLM network, each layer receives representations from all neurons in the previous layer, integrating and abstracting the previously learned local evaluation information to obtain a more comprehensive evaluation result.

Interestingly, our wider and deeper LLM network can be likened to the process of paper review. First, each reviewer independently assigns a score based on their own research background and understanding of the paper (the evaluation sample), representing the first layer. Then, a discussion phase follows, during which all reviewers take into account each other's evaluations to update their scores. This iterative process can continue through multiple rounds, analogous to subsequent layers in our network. Finally, the Chair or Editor consolidates all the reviewers' opinions to make the decision on whether the paper will be accepted.
The final experiments reveal that a LLM network with a wider scope yet limited to only two layers performs the best. This coincidence aligns with the current mainstream conference paper review process, where many reviewers are brought in for blind reviews and a single round of discussion, after which the chair makes the final decision.

% expert group evaluation process which involves a panel of experts who assess the quality of generated text. In the blind review process, experts independently evaluate the text quality from their own perspectives. And then the expert panel typically engages in discussions to reach a consensus on the evaluation. During these discussions, the experts may share their individual evaluations and refer to the opinions of other colleagues. The chair of the expert group may finally summarize the individual evaluations and feedback provided by the experts to arrive at a comprehensive assessment.

To facilitate the research on LLM evaluator, we also build a comprehensive benchmark that encompasses 15 tasks, such as question answering, text summarization, and programming. Additionally, the benchmark assesses 8 different abilities, such as logical reasoning, semantic understanding and text composition. To ensure thorough evaluation, we have compiled 2,553 samples, each of which comes with human-annotated preferences, 31 times larger than the dataset used in FairEval~\citep{wang2023large}.

The major contributions of this paper are summarized as follows:
\begin{itemize}
\item We explore the multi-layer wide network where each neuron possesses distinct neuron role and cooperative evaluations are performed among different layers of neurons. We observe that a wider two-layer LLM network, namely WideDeep, can achieve the best evaluation results, which is essentially a paper review process.

\item We introduce the largest and most diverse benchmark LLMEval$^2$ for LLM evaluator. LLMEval$^2$ involves diverse ability evaluation, and contributes to more sufficient assessment.

\item Our WideDeep network's effectiveness has been extensively validated through thorough experimentation on existing two benchmarks and LLMEval$^2$. This validation reveals a notable 3.5-point increase in accuracy, coupled with a noteworthy enhancement of 0.06 in the kappa correlation coefficient. Notably, we've successfully addressed a limitation previously identified in FairEval, where employing more than three LLMs failed to yield performance enhancements. This accomplishment underscores that augmenting the number of LLM neurons contributes to a more equitable evaluation process.

\item We also leverage WideDeep to assess the performance of the Chinese LLMs. WideDeep's advantages have further expanded compared to English benchmarks, with improvements of 6pts, 5.5pts, and 0.09 in accuracy, F1 score, and kappa correlation coefficient, respectively, achieving a labeling accuracy of 74\% and reaching a 93\% agreement level among humans. We demonstrate WideDeep has accelerated the LLM evaluation process by 4.6 times and decreased the average annotation cost per sample by 60\%.

\end{itemize}

\section{Related Work}

There has been a proliferation of LLM-based chatbots that harness instruction fine-tuning and learn from human feedback to unlock the ability of responding to questions following human preferences~\citep{bai2022training,yuan2023rrhf,song2023preference}.
However, assessing whether LLM is well aligned with human preference is not a straightforward task. 
Traditional LLM benchmarks like MMLU~\citep{hendrycks2020measuring} fall short in effectively distinguishing between these aligned models and the base models, as they only require LLM to answer multiple-choice questions.
Even if we have evaluation benchmarks available, such as several questions and manually annotated responses, commonly used ngram-based metrics like BLEU~\citep{papineni-etal-2002-bleu} and ROUGE~\citep{lin-2004-rouge}, as well as embedding-based metrics like BERTScore~\citep{zhang2019bertscore} and MoverScore~\citep{zhao2019moverscore}, can only measure lexical and semantic similarity between a generated response and the reference response. 
These metrics have been shown to have relatively low correlation with human judgments~\citep{liu2023gpteval}.

% \citeauthor{liu2023gpteval}

In recent research, it has been noticed that extensive generative pre-training has enabled LLMs to excel in assigning higher probabilities to high-quality responses based on given instructions and context~\citep{fu2023gptscore}. 
Building on this insight, researchers have leveraged ChatGPT and GPT-4 to evaluate numerous natural language generation tasks, including text summarization, story generation, data-to-text generation, and machine translation, showcasing remarkable performance~\citep{liu2023gpteval, wang2023chatgpt, kocmi2023large}. 
However, subsequent investigations have unveiled certain issues with LLM evaluators, particularly concerning biases related to position and verbosity~\citep{wang2023large, zheng2023judging}. To address these biases, researchers have adopted techniques such as swapping the order of candidate responses and conducting multiple independent evaluations, which effectively mitigates biases and yields more reliable results.
In this paper, we propose a unified approach, considering previous LLM evaluators as one-layer narrow LLM networks with varying numbers of neurons. Each neuron independently scores candidate samples from the same evaluation perspective. Drawing inspiration from deep neural networks, we delve into wider and deeper LLM networks, assigning distinct functionalities and roles to different LLM neurons. Each layer takes evaluation outputs from all neurons in the previous layer, resulting in a fairer LLM evaluator.
Furthermore, we contribute to the field by creating an extensive benchmark for evaluation across various tasks, aiming to drive progress and innovation in this research domain.

\section{Methodology}
% In this section, we provide a detailed description of our {\it Expert Group Evaluation Network} (EGE). We begin by introducing the process of assigning a perspective to each neuron in Sec.\ref{EPA}. Next, in Sec.\ref{MEN}, we outline the construction of a multi-layer evaluation network. Finally, we design aggregation strategies for the evaluation results in Sec.\ref{ERA}.
In this section, we begin by introducing the multi-layer wide LLM network in Sec.\ref{DWDN}. Next, we provide a more intuitive explanation from the perspective of academic paper review in Sec.\ref{EGR}.

\subsection{Deeper and Wider LLM Network}
\label{DWDN}
% Recently, researchers have utilized LLMs (e.g., GPT 3.5) to evaluate the quality of generated text, which create the prompt with two candidate responses to be evaluated. Afterwards the prompt is used to query the LLM for achieving the evaluation result. Representative methods, such as FairEval~\citep{wang2023large}, observe the positional bias and take a multiple evidence strategy that obtains the final evaluation score by ensembling the fixed number of evaluation results. They inquire the LLM before and after swapping the order of two responses appearing in the prompt context to obtain multiple evaluation scores, as shown in Figure~\ref{intro} (a). The evaluation procedure is essentially a single-layer neural network with a fixed number of neurons, in which LLM acts as the neuron and gives the evaluation score. Although impressive performance on benchmarks in terms of LLMs' ability to evaluation has been shown, it is still not clear whether LLMs can be further improved in deeper and wider network architecture.

State-of-the-art deep neural networks are composed of interconnected layers of neurons, where each neuron performs a specific function by processing input from other neurons and producing output for the next layer. 
At the bottom layer of the network, a considerable number of neurons are responsible for processing the input data and extracting diverse features that are relevant to the task at hand. 
As we move up the layers of the network, the neurons capture higher-level features and relationships by combining the lower-level features learned in preceding layers, which can be critical for solving more complex tasks. 
However, it remains unexplored whether widening and deepening the single-layer LLM network with a fixed number of neurons in Figure~\ref{intro} (a) can improve the evaluation performance. 
Inspired by this, we enhance the network by augmenting the number of neurons in each layer and increasing the depth of the network in Figure~\ref{intro} (b), making the LLM network deeper and wider.
Building such a network involves three key points: \textbf{The role of each neuron}, \textbf{The connection of different layers} and \textbf{The aggregation of final results}.

\textbf{The role of each neuron.} In deep neural networks, different neurons perform distinct functions where they may learn to respond to different linguistic features such as word order, grammar or semantics by back-propagation optimization. 
The role of each neuron is learned by gradient back-propagation to adjuest the neuron parameters.
However, within our LLM network, each neuron represents a frozen LLM, and we are unable to adjust the parameters of the network.
To keep different functions for LLM neurons, we first query LLMs to generate diverse neuron roles for each sample according to its content. 
Concretely, given a testing question $q$, two candidate responses $A=\{a_1, a_2\}$, a prompt ${\pi}_0$, and a template $\mathtt{F()}$, the generation of neuron roles describes a probability distribution $p_{\mathtt{LLM}}(\mathsf{P}|\mathtt{F}(q, A, {\pi}_0))$ over output perspectives $\mathsf{P}=\{\mathsf{p}_1, \mathsf{p}_2, ..., \mathsf{p}_n\}$ as computed by the $\mathtt{LLM}$. $\mathtt{F()}$ aims to fill the question $q$ and responses $A$ into the slots of prompt ${\pi}_0$. The neuron role prompt ${\pi}_0$ is summarized as follows:

\begin{thmblock}[Neuron Role Prompt ${\pi}_0$]
Please help me summarize that for a user question ``\{\{\textbf{question}\}\}'', if I want to determine which of two answers is better, from what angles do we need to evaluate? The two answers are respectively ``\{\{\textbf{answer\_1}\}\}'' and ``\{\{\textbf{answer\_2}\}\}''. 

Output the name and evaluation content of each angle. Each line is an evaluation angle. Use a newline to separate different evaluation angles. Each evaluation angle Name starts with \$ and ends with \&.
\end{thmblock}
% \begin{tcolorbox}[colback=black!3!white,colframe=black!16!white]
% Please help me summarize that for a user question ``\{\{\textbf{question}\}\}'', if I want to determine which of two answers is better, from what angles do we need to evaluate? The two answers are respectively ``\{\{\textbf{answer\_1}\}\}'' and ``\{\{\textbf{answer\_2}\}\}''. 

% Output the name and evaluation content of each angle. Each line is an evaluation angle. Use a newline to separate different evaluation angles. Each evaluation angle Name starts with \$ and ends with \&.
% \end{tcolorbox}

For the generated neuron roles $\mathsf{P}=\{\mathsf{p}_1, \mathsf{p}_2, ..., \mathsf{p}_n\}$, we respectively assign $\mathsf{p}_i$ to each neuron $\mathsf{n}_i$ in each layer, simulating the different roles of neurons in deep neural networks. For example, as shown in Figure~\ref{intro} (b), the LLM, such as $\mathtt{gpt}$-$\mathtt{3.5}$-$\mathtt{turbo}$, generates four perspectives including coherence, relevance, harmlessness and accuracy, and then the LLM network would possess four neurons in each layer where each neuron played by the LLM is respectively responsible for evaluating the candidate responses from one of the four perspectives. 
For the input layer of LLM network, given a prompt ${\pi}_1$, and a template $\mathtt{F()}$, each neuron $\mathsf{n}_i$ defines a probability distribution $p_{\mathtt{LLM}}^i(\mathsf{e}_1^i|q, A)$ over output evaluation result $\mathsf{e}_1^i$ as computed by the $\mathtt{LLM}$:
\begin{align}
p_{\mathtt{LLM}}^i(\mathsf{e}_1^i|q, A) = p_{\mathtt{LLM}}^i(\mathsf{e}_1^i|\mathtt{F}(q, A, \mathsf{p}_i,{\pi}_1))p_{\mathtt{LLM}}(\mathsf{p}_i|\mathtt{F}(q, A, {\pi}_0))
\label{eq1}
\end{align}
% $\mathtt{F()}$ aims to fill the question $q$, answers $A$ and perspective $\mathsf{p}_i$ into the slots of prompt ${\pi}_1$. 
where the input layer evaluation prompt ${\pi}_1$ for LLMs is described as follows:

\begin{colblock}[Input Layer Evaluation Prompt ${\pi}_1$]
You are a member of the expert group for checking the quality of answer. You are given a question and two answers. Your job is to decide which answer is better for replying question.

[Question]

\{\{\textbf{question}\}\}

[The Start of Assistant 1's Answer]

\{\{\textbf{answer\_1}\}\}

[The End of Assistant 1's Answer]

[The Start of Assistant 2's Answer]

\{\{\textbf{answer\_2}\}\}

[The End of Assistant 2's Answer]

[System]

Take \{\{\textbf{perspective}\}\} as the Angle of View, we would like to request your feedback on the performance of two AI assistants in response to the user question displayed above.

Each assistant receives an overall score on a scale of 1 to 10, ...

...

PLEASE OUTPUT WITH THE FOLLOWING FORMAT:

<start output>

Evaluation evidence: <your evaluation explanation here>

Score of Assistant 1: <score>

Score of Assistant 2: <score>

<end output>
    
Now, start your evaluation:
\end{colblock}
% \begin{tcolorbox}[colback=black!3!white,colframe=black!16!white]
% You are a member of the expert group for checking the quality of answer. You are given a question and two answers. Your job is to decide which answer is better for replying question.

% [Question]

% \{\{\textbf{question}\}\}

% [The Start of Assistant 1's Answer]

% \{\{\textbf{answer\_1}\}\}

% [The End of Assistant 1's Answer]

% [The Start of Assistant 2's Answer]

% \{\{\textbf{answer\_2}\}\}

% [The End of Assistant 2's Answer]

% [System]

% Take \{\{\textbf{perspective}\}\} as the Angle of View, we would like to request your feedback on the performance of two AI assistants in response to the user question displayed above.

% Each assistant receives an overall score on a scale of 1 to 10, ...

% ...

% PLEASE OUTPUT WITH THE FOLLOWING FORMAT:

% <start output>

% Evaluation evidence: <your evaluation explanation here>

% Score of Assistant 1: <score>

% Score of Assistant 2: <score>

% <end output>
    
% Now, start your evaluation:
% \end{tcolorbox}

\textbf{The connection of different layers.} In naive deep neural networks, the neurons in each layer are interconnected through weighted connections. These connections are responsible for transmitting information from one layer to the next during the forward pass of the network. Concretely, within each hidden layer, each neuron is connected to all the neurons in the previous layer. The connections between neurons in the hidden layers are weighted, and the weights are learned through the training process to allow the network to capture and represent complex patterns and features from the input data. 
In our LLM network, there is neither numerical weights nor training optimization. Therefore, inspired by Stacked LLMs~\citep{sordoni2023deep}, we write the prompt ${\pi}_2$ which serves as the weights to connect each neuron with all neurons in the previous layer. 
Similarly, each neuron $\tilde{\mathsf{n}}_i$ in the $l_{th}$ layer defines a probability distribution $p_{\mathtt{LLM}}^i(\mathsf{e}_l^i|q, A)$ over output evaluation result $\mathsf{e}_l^i$ as computed by the $\mathtt{LLM}$:
% $\mathtt{F()}$ aims to fill the question $q$, answers $A$ and perspective $\mathsf{p}_i$ into the slots of prompt ${\pi}_1$. The input layer evaluation prompt ${\pi}_1$ for LLMs is described as follows:
% The query prompt equivalent to the transformed information from previous layer to the current is written as follows:

\begin{align}
p_{\mathtt{LLM}}^i(\mathsf{e}_l^i|q, A) = \sum_{j=1}^n p_{\mathtt{LLM}}^i(\mathsf{e}_l^i|\mathtt{F}(q, A, \mathsf{e}_{l-1}^j, \mathsf{p}_{l-1}^j, {\pi}_2))p_{\mathtt{LLM}}^j(\mathsf{e}_{l-1}^j|\mathtt{F}(q, A, \mathsf{p}_j, {\pi}_1))
\label{eq2}
\end{align}
where $n$ is the number of neurons in the previous layer, $\mathsf{p}_{l-1}^j$ is the role of $j_{th}$ neuron in the $(l-1)_{th}$ layer. ${\pi}_2$ is the hidden layer evaluation prompt for LLMs which is described as follows:

\begin{propblock}[Hidden Layer Evaluation Prompt ${\pi}_2$]
You are a member of the expert group for checking the quality of answer. You are given a question and two answers. Your job is to decide which answer is better for replying question.

[Question]

\{\{\textbf{question}\}\}

[The Start of Assistant 1's Answer]

\{\{\textbf{answer\_1}\}\}

[The End of Assistant 1's Answer]

[The Start of Assistant 2's Answer]

\{\{\textbf{answer\_2}\}\}

[The End of Assistant 2's Answer]

[System]

You and your colleagues in the expert group have conducted several rounds of evaluations.

[The Start of Your Historical Evaluations]
    
\{\{\textbf{Your own evaluation from last layer}\}\}
    
[The End of Your Historical Evaluations]

[The Start of Other Colleagues' Evaluations]
    
\{\{\textbf{Other evaluations from last layer}\}\}

[The End of Other Colleagues' Evaluations]
    
Again, take \{\{\textbf{inherited perspectives}\}\} as the Angle of View, we would like to request your feedback on the performance of two AI assistants in response to the user question displayed above.
Each assistant receives an overall score on a scale of 1 to 10, ...

...

PLEASE OUTPUT WITH THE FOLLOWING FORMAT:

<start output>

Evaluation evidence: <your evaluation explanation here>

Score of Assistant 1: <score>

Score of Assistant 2: <score>

<end output>
    
Now, start your evaluation:
\end{propblock}
% \begin{tcolorbox}[colback=black!3!white,colframe=black!16!white,breakable]
% You are a member of the expert group for checking the quality of answer. You are given a question and two answers. Your job is to decide which answer is better for replying question.

% [Question]

% \{\{\textbf{question}\}\}

% [The Start of Assistant 1's Answer]

% \{\{\textbf{answer\_1}\}\}

% [The End of Assistant 1's Answer]

% [The Start of Assistant 2's Answer]

% \{\{\textbf{answer\_2}\}\}

% [The End of Assistant 2's Answer]

% [System]

% You and your colleagues in the expert group have conducted several rounds of evaluations.

% [The Start of Your Historical Evaluations]
    
% \{\{\textbf{Your own evaluation from last layer}\}\}
    
% [The End of Your Historical Evaluations]

% [The Start of Other Colleagues' Evaluations]
    
% \{\{\textbf{Other evaluations from last layer}\}\}

% [The End of Other Colleagues' Evaluations]
    
% Again, take \{\{\textbf{inherited perspectives}\}\} as the Angle of View, we would like to request your feedback on the performance of two AI assistants in response to the user question displayed above.
% Each assistant receives an overall score on a scale of 1 to 10, ...

% ...

% PLEASE OUTPUT WITH THE FOLLOWING FORMAT:

% <start output>

% Evaluation evidence: <your evaluation explanation here>

% Score of Assistant 1: <score>

% Score of Assistant 2: <score>

% <end output>
    
% Now, start your evaluation:
% \end{tcolorbox}
{\it \textbf{Note:}} The slot ``\{\{\textbf{Your own evaluation from last layer}\}\}'' should be filled in the output evaluation evidece and score of the neuron in the prior layer that corresponds to the same position as the current neuron, while other neurons' output values are filled in ``\{\{\textbf{Other evaluations from last layer}\}\}''. The slot ``\{\{\textbf{inherited perspectives}\}\}'' represents the union of all neurons' roles in the previous layer.

\textbf{The aggregation of final results.} The output layer in deep neural network generates the final prediction for the task. Similarly, we aggregate the evaluation results from neurons in the LLM network, and there actually exists a variety of aggregation strategies to derive the ultimate evaluation conclusion. The strategies employed in this study include: (1) Averaging the scores from all neurons in the network for each response and subsequently comparing the average scores of the responses to determine which is better (${\mathtt{c}}^*_1$). (2) Comparing the evaluation scores of the responses from each neuron to choose the better one, and then voting over neurons in all layers or each layer (${\mathtt{c}}^*_2$).
\begin{equation}
\begin{aligned}
{\mathtt{c}}^*_1 = \mathtt{argmax} \sum_{l} \sum_{i=1}^n p_{\mathtt{LLM}}^i(\mathsf{e}_l^i|q, A) \\
{\mathtt{c}}^*_2 = \mathtt{max}_{\mathtt{count}} \bigcup_{i,l}\{\mathtt{argmax} \, p_{\mathtt{LLM}}^i(\mathsf{e}_l^i|q, A)\}
\label{eq3}
\end{aligned}
\end{equation}

\begin{figure}[t]
\centering
\includegraphics[width=1.0\columnwidth]{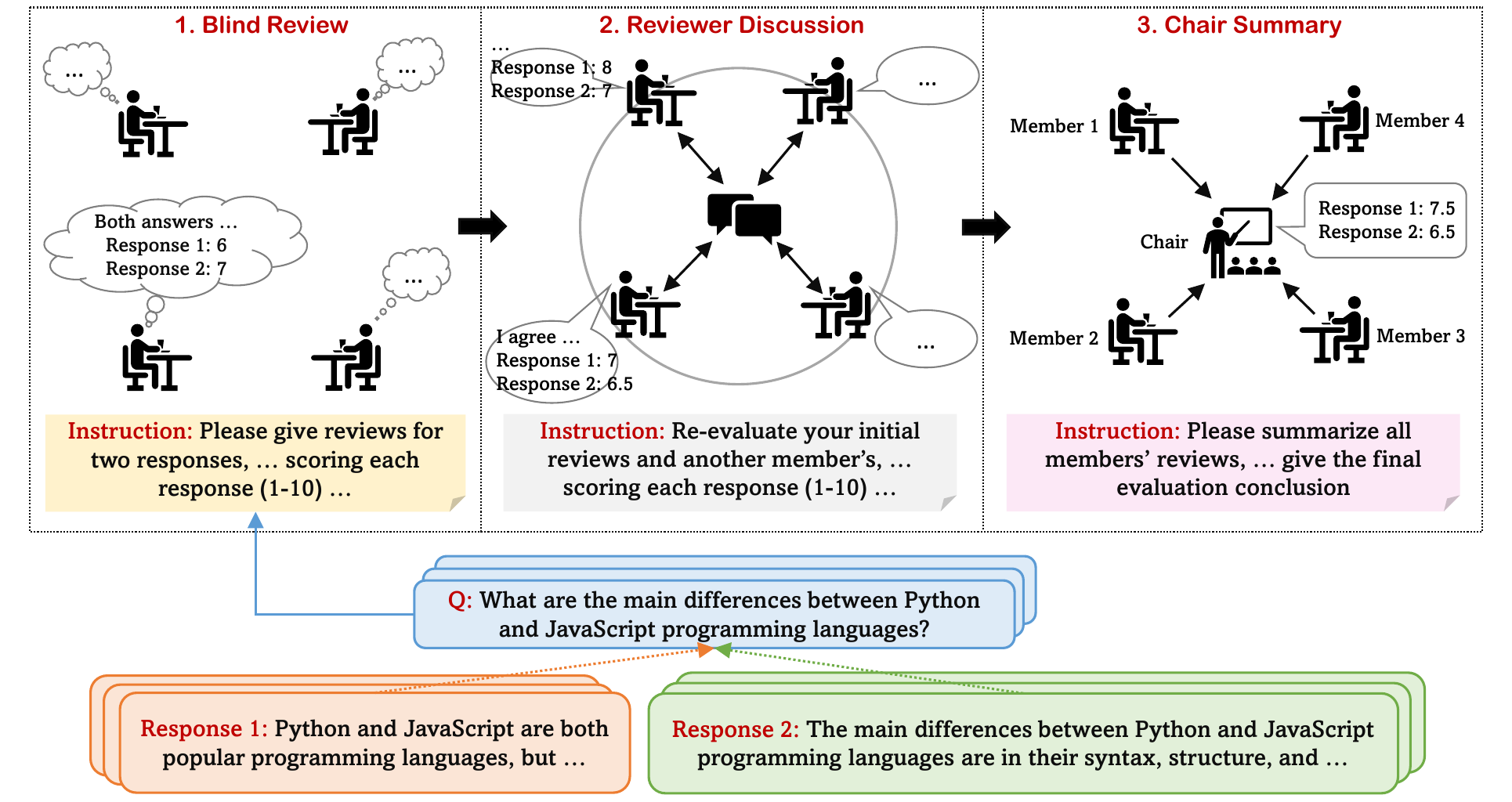} % Reduce the figure size so that it is slightly narrower than the column. Don't use precise values for figure width.This setup will avoid overfull boxes.
\caption{Academic paper review process for evaluating the quality of candidate responses, comprising of blind review, reviewer discussion and chair summary.}
\label{expg}
\end{figure}

\subsection{Explain LLM Network as a Academic Paper Review Process}
\label{EGR}
To offer more coherent insights into our deeper and wider LLM network, we present an analogy using the perspective of academic paper review, as depicted in Figure~\ref{expg}. 
The review process usually consists of three key stages: {\it blind review}, {\it reviewer discussion}, and {\it chair summary}.
In the blind review stage, each reviewer diligently examines the candidate paper based on their research background. 
Subsequently, they provide feedback in the form of a written report, akin to the input layer of our LLM network.
Following the individual blind reviews, reviewers may engage in discussions to further evaluate the candidate responses. 
These discussions resemble the subsequent layers of our network, where reviewers compare and contrast their assessments, explore areas of agreement or disagreement, and identify potential biases or blind spots. 
This iterative process of discussion can span multiple rounds, analogous to the deep layers in our network.
Finally, the chair makes a decision, akin to the result aggregation step in our network, by considering the collective feedback from the reviewers.
By illustrating the functioning of our LLM network through the academic paper review analogy, we aim to provide a more intuitive understanding of its operations and effectiveness.

% \subsection{Evaluation Perspective Assignment}
% \label{EPA}
% Evaluating the quality of generated text is subjective and reference-free, which is imperative to assess from a multitude of dimensions. To achieve this, we utilize LLMs (e.g., ChatGPT) to generate diverse evaluation perspectives for each evaluation sample according to its content.
% \begin{tcolorbox}[colback=black!3!white,colframe=black!16!white]
% Please help me summarize that for a user question ``\{\{\textbf{question}\}\}'', if I want to determine which of two answers is better, from what angles do we need to evaluate? The two answers are respectively ``\{\{\textbf{answer\_1}\}\}'' and ``\{\{\textbf{answer\_2}\}\}''. 

% Output the name and evaluation content of each angle. Each line is an evaluation angle. Use a newline to separate different evaluation angles. Each evaluation angle Name starts with \$ and ends with \&.
% \end{tcolorbox}

% \subsection{Multi-layer Evaluation Network}
% \label{MEN}

% \subsection{Evaluation Result Aggregation}
% \label{ERA}

\begin{figure}[t]
\centering
\includegraphics[width=1.0\columnwidth]{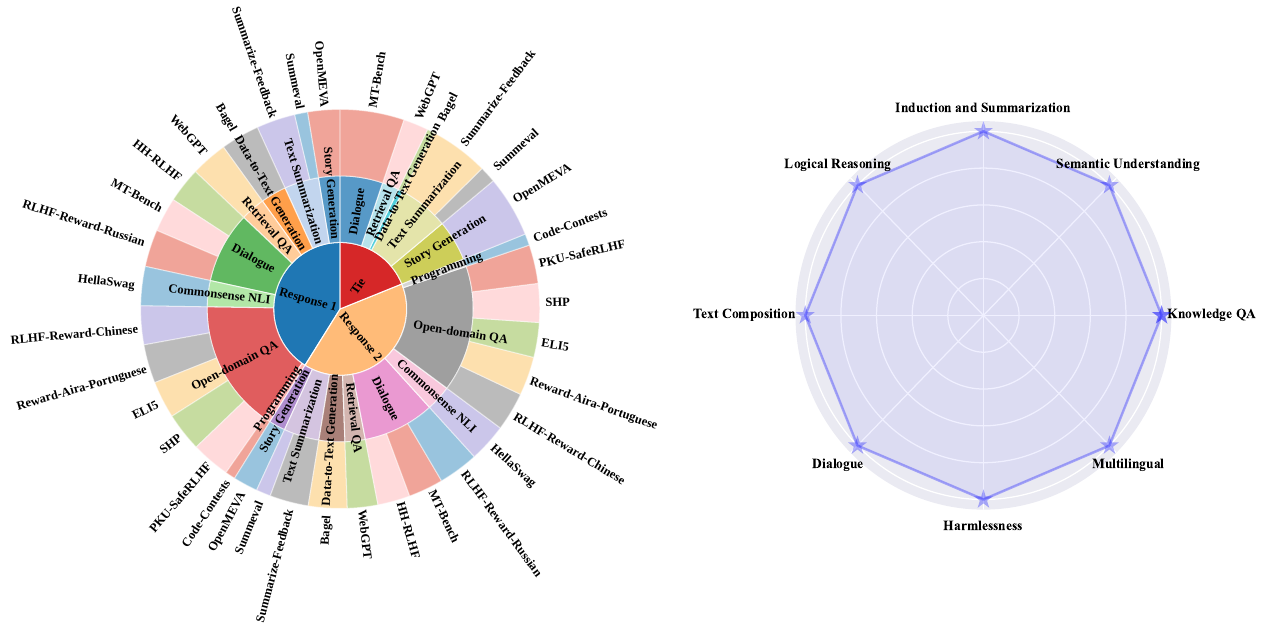} 
% Reduce the figure size so that it is slightly narrower than the column. Don't use precise values for figure width.This setup will avoid overfull boxes.
\caption{{\it \textbf{Left}} is the distribution of all datasets in LLMEval$^2$. The outer and middle circles display the names of datasets and their associated tasks, respectively. The inner circle represents the proportions of three categories of data in the benchmark concerning the preference between two responses: the first one being better, the second one being better, or the two responses having similar quality. {\it \textbf{Right}} illustrates covered 8 evaluation abilities of LLMEval$^2$.}
\label{benchmark}
\end{figure}

% Please read the instructions below carefully and follow them faithfully. \textbf{Important:} This year the checklist will be submitted separately from the main paper in OpenReview, please review it well ahead of the submission deadline: \url{https://neurips.cc/public/guides/PaperChecklist}.

\section{LLMEval$^2$ Benchmark}

In addition to exploring the wider and deeper LLM network to obtain fairer evaluation results, we also seek to propose improvements to the current LLM evaluator benchmark. 
The widely used benchmarks, such as FairEval~\citep{wang2023large} and MT-bench~\citep{zheng2023judging}, only consist of 80 testing samples, leading to unstable evaluation results and making it challenging to comprehensively assess the LLM evaluator's capabilities.
While PandaLM constructes a test set comprising 999 samples, it still lacks statistics for different abilities and suffers from a limitation in data diversity, as it solely relies on a single self-instruct source~\citep{wang-etal-2023-self-instruct}.
To address these shortcomings, we present LLMEval$^2$, the largest and most diverse evaluation benchmark for the LLM Evaluator to date.

\textbf{Benchmark Construction.} Assessing the capabilities of the LLM evaluator requires data that includes a question, a pair of candidate responses, and a human label indicating the preferred response. 
We notice that the format of the evaluation data resembles that of the samples used to train a reward model. The reward trainer aims to grasp human preferences by ranking the candidate responses based on human labels. Thus, we  compile datasets used for training a reward model, totaling 15 datasets (shown as the outer circle in Figure~\ref{benchmark} {\it \textbf{left}}). Next, we employ data sampling techniques to balance data diversity and evaluation costs, resulting in a collection of 2,553 evaluation samples, each annotated with human preferences, across all 15 datasets.

\textbf{Statistics.} In this benchmark, 1,050 samples of response 1 are considered to align with human preferences, while another 1,021 samples of response 2 are deemed superior. Additionally, two responses from the 482 samples are considered difficult to differentiate in terms of quality. 
As illustrated in Figure~\ref{benchmark} (\textit{\textbf{left}}), the benchmark encompasses eight tasks: Story Generation, Text Summarization, Data-to-Text Generation, Retrieval QA, Dialogue, Commonsense NLI, Open-domain QA, and Programming. These tasks evaluate eight abilities of the benchmark: {\it Induction and Summarization}, {\it Semantic Understanding}, {\it Knowledge QA}, {\it Logical Reasoning}, {\it Text Composition}, {\it Dialogue}, {\it Harmlessness} and {\it Multilingual}.

% The benchmark 中 1000条样本的response 1被认为符合人类偏好，另外1000条样本的response 2被认为更好，还有500条样本中的两个response被认为难以区分出质量差异. As shown in Figure~\ref{benchmark} ({\it \textbf{left}}), eight tasks are contained, {\it Story Generation}, {\it Text Summarization}, {\it Data-to-Text Generation}, {\it Retrieval QA}, {\it Dialogue}, {\it Commonsense NLI}, {\it Open-domain QA} and {\it Programming}. Among them, the benchmark  holds eight abilities, including {\it Induction and Summarization}, {\it Semantic Understanding}, {\it Knowledge QA}, {\it Logical Reasoning}, {\it Text Composition}, {\it Dialogue}, {\it Harmlessness} and {\it Multilingual}.

% To the best of our knowledge, this benchmark currently represents the largest scale for evaluating the performance of LLMs on assessing the quality of generated text, and it possesses more diverse tasks and evaluation abilities.

\section{Experiments}
In this section, our primary focus is to address the following research questions: (\textbf{RQ1}) Does a LLM network with a wider and deeper structure yield improved evaluation performance? (\textbf{RQ2}) Which neuron roles does LLM prioritize, and how do they impact the results? (\textbf{RQ3}) To what extent can our LLM evaluator accelerate manual annotation speed in real LLM business?

\subsection{Experimental Settings}
\textbf{Datasets.} We conduct evaluations on three benchmarks, consisting of two existing datasets, FairEval~\citep{wang2023large} and PandaLM~\citep{wang2023pandalm}, along with our newly constructed dataset, LLMEval$^2$. FairEval comprises a total of 80 samples, and the candidate responses are generated by Vicuna-13b and ChatGPT. Meanwhile, PandaLM consists of 999 samples, which were drawn from the diverse human evaluation dataset of self-instruct~\citep{wang-etal-2023-self-instruct}. The paired responses in PandaLM are generated by LLaMA-7B, Bloom-7B, Cerebras-GPT-6.7B, OPT-7B, and Pythia-6.9B.

\textbf{Implementation Details.} We use accuracy (Acc), Macro-F1, and the kappa correlation coefficient (Kap.) as our evaluation metrics. For reporting the main results, we utilize $\mathtt{gpt}$-$\mathtt{3.5}$-$\mathtt{turbo}$ as the LLM neuron on the full dataset due to cost constraints. 
Additionally, we construct a smaller version called LLMEval$^2$ mini, which consists of 20 samples drawn from each of the 15 datasets, resulting in a total of 300 samples. 
These samples are used for analytical experiments.
% , and we further validate the effectiveness of our network in conjunction with the more powerful $\mathtt{gpt}$-$\mathtt{4}$ as the LLM neuron.

\subsection{Experimental Results}
Table~\ref{tb1} shows the main results of our multi-layer wide LLM network $\mathtt{WideDeep}$ compared with prior single-layer network with fixed number of neurons $\mathtt{FairEval}$~\citep{wang2023large}. 
We implement four variants $\mathtt{WideDeep}$ $ {\mathtt{c}}^*_1$, $\mathtt{WideDeep}$ $ {\mathtt{c}}^*_2 (l_1)$, $\mathtt{WideDeep}$ $ {\mathtt{c}}^*_2 (l_2)$ and $\mathtt{WideDeep}$ $ {\mathtt{c}}^*_2 (all)$. $\mathtt{WideDeep}$ $ {\mathtt{c}}^*_1$ indicates averaging the scores from all neurons in LLM network and choosing the response with higher score (${\mathtt{c}}^*_1$ in Equation~\ref{eq3}). For the latter three, we aggregate the results based on ${\mathtt{c}}^*_2$ in Equation~\ref{eq3}. $\mathtt{WideDeep}{\mathtt{c}}^*_2 (l_1)$ represents voting the evaluation results only in the $1_{st}$ layer and $\mathtt{WideDeep}$ $\mathtt{W} \, {\mathtt{c}}^*_2 (l_2)$ means only voting in the $2_{nd}$ layer of LLM network. Voting all evaluation results in all layers is denoted as $\mathtt{WideDeep}$ $ {\mathtt{c}}^*_2 (all)$. 
The best results over evaluation metrics are in bold.
Note that we have attempted to use deeper LLM networks (more than 2 layers), but it resulted in a decrease in performance. Therefore, in our main experiment, we do not restrict the number of neurons in each layer, but we limit the network depth to 2 layers. We will discuss the impact of network depth on the results in the analysis experiment.

We can observe that our multi-layer wide LLM network outperforms FairEval significantly, with an increase in accuracy by 3.2pts, 4.4pts, and 3pts, and an improvement in kappa correlation by 3.7pts, 8.4pts, and 6.3pts on the three respective benchmarks.
 Compared with voting in each layer of the LLM network $\mathtt{WideDeep}$ $ {\mathtt{c}}^*_2 (l_1)$ and $\mathtt{WideDeep}$ $ {\mathtt{c}}^*_2 (l_2)$, $\mathtt{WideDeep}$ $ {\mathtt{c}}^*_2 (all)$ which votes evaluation results from all layers achieves the better overall performance. Meanwhile, in comparison with$\mathtt{WideDeep}$ $ {\mathtt{c}}^*_2 (l_1)$, $\mathtt{WideDeep}$ $ {\mathtt{c}}^*_2 (l_2)$ reaches the higher performance which demonstrates that the effectiveness of deepening the LLM network.

 \begin{table}[t]
\renewcommand\arraystretch{1.2}
\tabcolsep=0.13cm
\centering
\small
% \begin{adjustbox}{angle=90}
\caption{Main Results on FairEval, PandaLM and LLMEval$^2$ benchmarks.}
\label{tb1}
\begin{tabular}{cccc|ccc|ccc}
\toprule
&\multicolumn{3}{c|}{\textbf{FairEval Benchmark}} &\multicolumn{3}{c|}{\textbf{PandaLM Benchmark}} &\multicolumn{3}{c}{\textbf{LLMEval$^2$ Benchmark}} \\
&Acc & Macro-F1 & Kap. &Acc & Macro-F1 & Kap. &Acc & Macro-F1 & Kap. \\ \midrule
FairEval~\citep{wang2023large} &0.587 & -- &0.31 &0.7147 &0.5531 &0.4891 &0.5735 &0.4663 & 0.2807 \\ \midrule
$\mathtt{WideDeep}$ $ {\mathtt{c}}^*_1$ &0.6063 & 0.4457 &0.3336 &0.7447 &0.5834 &0.5371 &0.5946 &0.4446 & 0.3197 \\ 
$\mathtt{WideDeep}$ $ {\mathtt{c}}^*_2 (l_1)$ &0.6125 & 0.4394 &0.3215 &0.7467 &0.6481 &0.5524 &0.5895 &0.4622 & 0.3155 \\
$\mathtt{WideDeep}$ $ {\mathtt{c}}^*_2 (l_2)$ &\textbf{0.6188} & \textbf{0.4479} &\textbf{0.3472} &0.7447 &0.6295 &0.5504 &0.5962 &0.5028 & 0.3345 \\
$\mathtt{WideDeep}$ $ {\mathtt{c}}^*_2 (all)$ &\textbf{0.6188} & 0.4465 &0.3462 &\textbf{0.7568} &\textbf{0.6545} & \textbf{0.5726} &\textbf{0.6036} &\textbf{0.5041} & \textbf{0.3440} \\
\bottomrule
\end{tabular}
\end{table}

\begin{figure}[t]
\centering
\includegraphics[width=0.7\columnwidth]{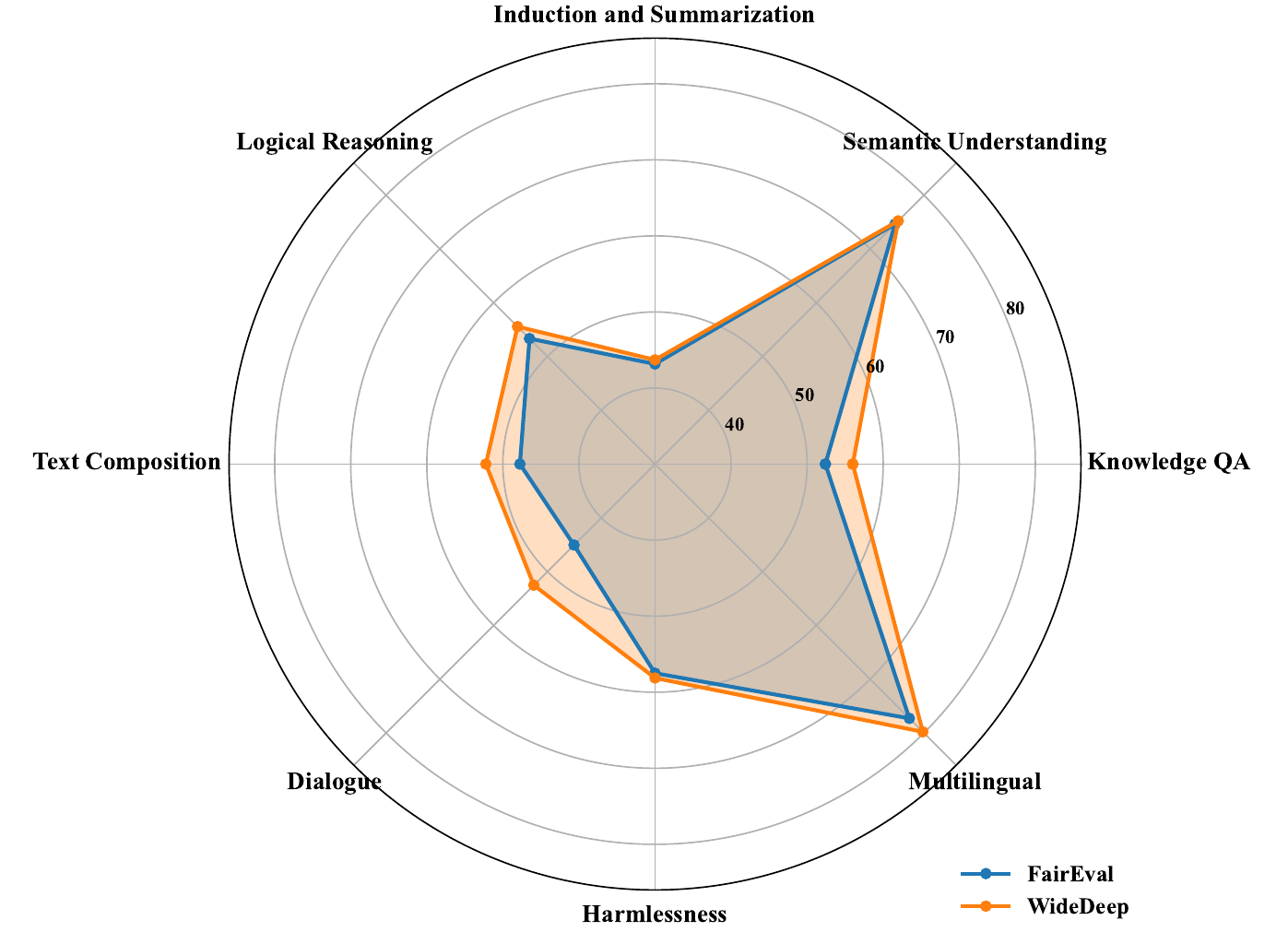} % Reduce the figure size so that it is slightly narrower than the column. Don't use precise values for figure width.This setup will avoid overfull boxes.
\caption{Comparison of accuracy between WideDeep and FairEval under eight abilities.}
\label{compare-ability}
\end{figure}

\subsection{Experimental Analyses}

Due to cost constraints, we extract 20 samples from each of the 15 datasets included in LLMEval$^2$, resulting in a total of 300 testing samples, namely LLMEval$^2$ mini. This mini dataset allows us to easily assess the impact of network width, depth and neuron roles.
% , and the use of GPT-4 as an LLM neuron on the results.

\begin{table}[t]
\renewcommand\arraystretch{1.2}
\centering
% \small
% \begin{adjustbox}{angle=90}
\caption{Performance on wider and deeper network. $\mathtt{NL}$ indicates no limit on the number of neurons.}
\label{tb3}
\begin{tabular}{ccccccc}
\toprule
&&$n=1$ & $n=2$ &$n=3$ & $n=4$ &$n=\mathtt{NL}$\\ \midrule
\multirow{2}{*}{$l=1$} & Acc &0.6033 & 0.6333 &0.6300 & 0.6267 & 0.6300 \\ 
& Macro-F1 &0.4709 & 0.4704 &0.4793 & 0.4885 &0.5116 \\ \midrule
\multirow{2}{*}{$l=2$} & Acc &0.6333 & 0.6400 &0.6433 & 0.6500 & 0.6567 \\ 
& Macro-F1 &0.4819 & 0.5187 &0.4772 & 0.5159 & 0.5666 \\ \midrule
\multirow{2}{*}{$l=3$} & Acc &0.6533 & 0.6400 &0.6433 & 0.6300 & 0.6500 \\ 
& Macro-F1 &0.5076 & 0.5084 &0.4764 & 0.4798 & 0.5053 \\
\bottomrule
\end{tabular}
\end{table}

\textbf{Wider LLM network is a Fairer Evaluator.}  Table~\ref{tb3} illustrates the performance improvement as the number of neurons in each layer of the LLM network ($n$) increases. When the number of layers $l$ is limited to one or two, we observe a consistent upward trend in performance. This demonstrates the effectiveness of widening the LLM network, fully unleashing the potential of a group of neurons.

\textbf{Slightly deeper LLM network is a Fairer Evaluator.}  From Table~\ref{tb3}, we can also observe that increasing the number of layers ($l$) in the network from 1 to 2 while keeping the number of neurons per layer fixed resulted in significant performance improvements. However, further deepening the network led to a slight decline in performance. The reason for this could be that deeper LLM networks tend to hold more homogeneous information, similar to overfitting in deep neural networks. 

\begin{figure}[t]
\centering
\includegraphics[width=0.86\columnwidth]{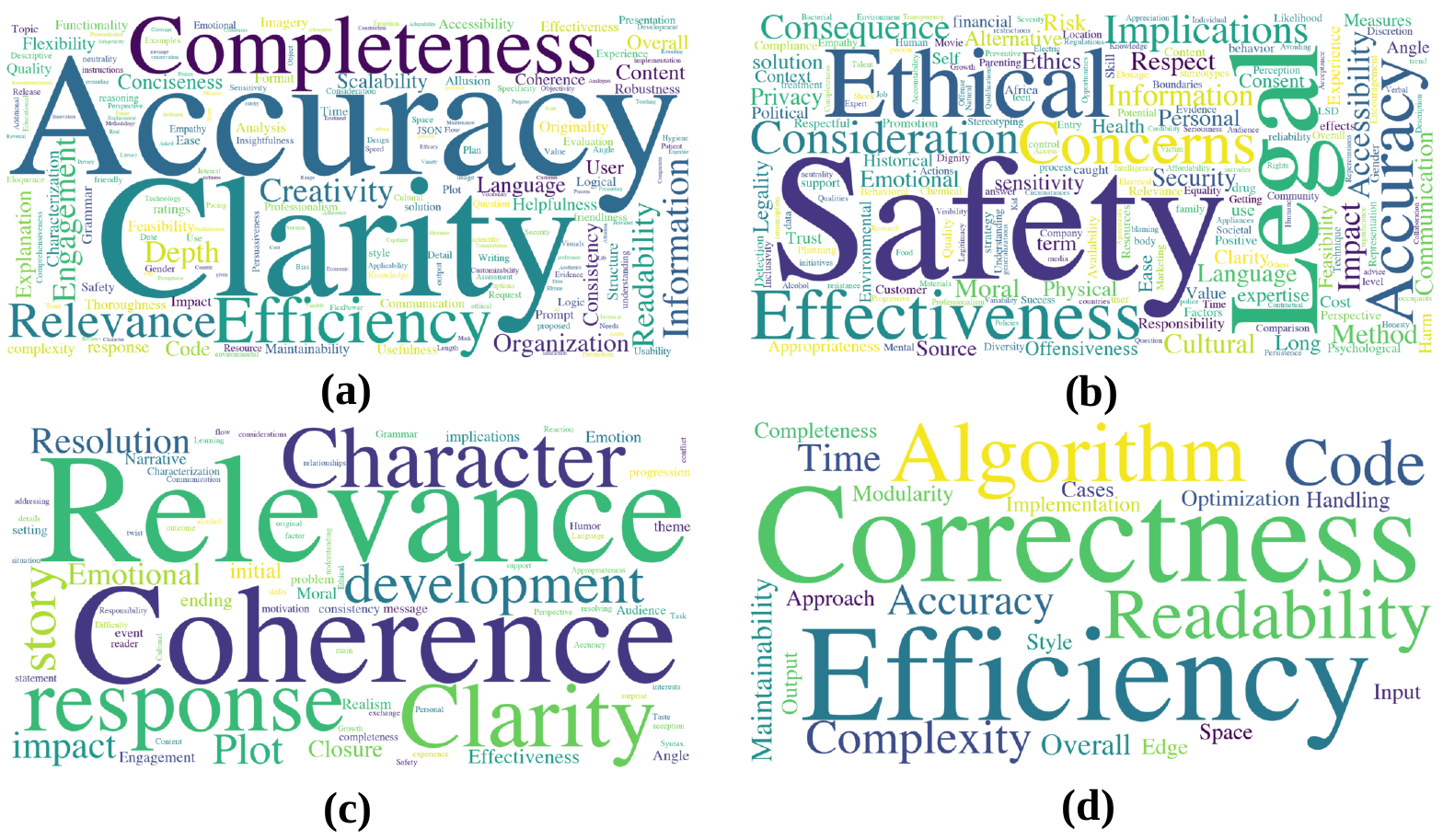} % Reduce the figure size so that it is slightly narrower than the column. Don't use precise values for figure width.This setup will avoid overfull boxes.
\caption{Word clouds of neuron roles on \textbf{(a)} {\it Dialogue} \textbf{(b)} {\it Harmlessness QA} \textbf{(c)} {\it Story Generation} \textbf{(d)} {\it Programming} task.}
\label{asp-visual}
\end{figure}

\begin{table}[tb]
\renewcommand\arraystretch{1.2}
\centering
% \small
% \begin{adjustbox}{angle=90}
\caption{Effectiveness of neuron roles. $\mathtt{NL}$ indicates no limit on the number of neurons in each layer.}
\label{tb2}
\begin{tabular}{ccc}
\toprule
&Acc & Macro-F1 \\ \midrule
WideDeep $(l=2, n=2)$ &0.6400 & 0.5187 \\ 
WideDeep $(l=2, n=2) \, \mathtt{W/O} \, \mathtt{Neuron} \, \mathtt{Roles}$ &0.6267 & 0.4992 \\ \midrule
WideDeep $(l=2, n=\mathtt{NL})$ &0.6567 & 0.5666 \\ 
WideDeep $(l=2, n=\mathtt{NL}) \, \mathtt{W/O} \, \mathtt{Neuron} \, \mathtt{Roles}$ &0.6400 & 0.5086 \\
\bottomrule
\end{tabular}
\end{table}

\textbf{Neuron roles are diverse and effective.} To mimic the characteristic of different neurons in a neural network being responsible for detecting different concepts, we require the LLM to generate potential evaluation dimensions before assessing the samples. In the network, each LLM in every layer is responsible for evaluating one specific dimension. 
To elucidate the roles that LLM assigns to neurons for each task, we present word clouds for four tasks in Figure~\ref{asp-visual}: dialogue, harmlessness QA, story generation, and programming. 
Note that we did not explicitly provide task names or definitions to LLM when generating the roles.
Remarkably, these assigned roles appear to be logical and adaptable, dynamically changing based on the specific task characteristics.
For harmlessness QA, LLM generates roles related to security, including $\mathsf{Safety}$, $\mathsf{Legal}$, and $\mathsf{Ethical}$. In  story generation, LLM assigns roles like $\mathsf{Coherence}$, $\mathsf{Relevance}$, and $\mathsf{Character}$. Meanwhile, the programming task involves algorithm-related roles, such as $\mathsf{Correctness}$ and $\mathsf{Efficiency}$. Having reliable and diverse neuron roles allows the LLM network to effectively utilize multiple neurons' value when the network becomes wider. As illustrated in Table~\ref{tb2}, we conduct two groups of experiments where the number of layers $l$ is set to 2 and neurons $n$ to no limit, respectively. The results show that the accuracy and Macro-F1 metrics decrease by 1.33\%, 1.67\% and 1.95\%, 5.80\% without neuron roles.

\begin{figure}[t]
\centering
\includegraphics[width=0.9\columnwidth]{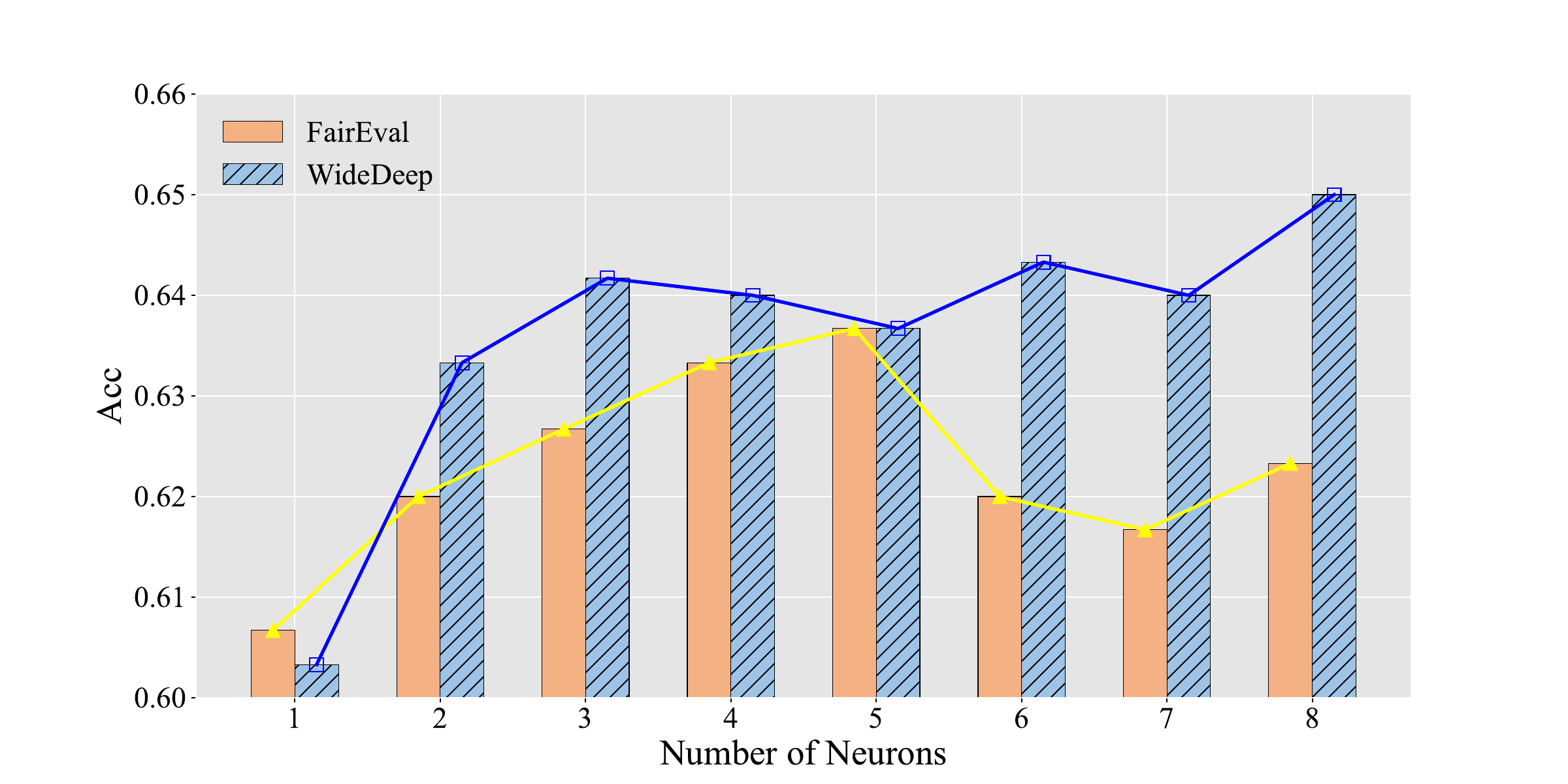} 
% Reduce the figure size so that it is slightly narrower than the column. Don't use precise values for figure width.This setup will avoid overfull boxes.
\caption{Performance under different neuron quantity constraints.}
\label{agent_num}
\end{figure}

\textbf{Widedeep can consume more neurons than baselines.} With a wider and deeper architecture and diverse neuron roles, our WideDeep network can utilize an unlimited number of LLM neurons. Previous methods, such as FairEval~\citep{wang2023large}, can also harness a large number of LLM neurons by integrating multiple independent LLM evaluations. 
In Figure~\ref{agent_num}, we demonstrate that Deepwide can more efficiently leverage LLM neurons to achieve significantly improved accuracy across almost all neuron quantity constraints than FairEval. 
Moreover, as the number of neurons increases, the performance continues to improve. 
For our experiments, we opted for a two-layered Deepwide network, where, with an odd-numbered neuron constraint, the second layer's neurons are reduced by one.
On the other hand, FairEval's performance saturates when the number of neurons reaches five, and any further increase leads to a decline in performance. This observation aligns with the conclusions of the original research, further confirming the positive impact of our deeper network and diversified neuron roles.

\subsection{Application in Chinese LLM Evaluation}

We also utilize WideDeep to assess the performance of the Chinese LLMs by determining which of the three responses under the same prompt is better.
Due to variations in evaluation data and tasks, the traditional manual annotation process involves multiple steps such as annotator training, small-scale trial annotation, selection of official annotators, and cross-annotation by multiple individuals. 
However, with the assistance of WideDeep, this process has been simplified to involve only a fixed team of professional annotators who perform sampling checks on the results generated by WideDeep.

\begin{table}[t]
\renewcommand\arraystretch{1.2}
\centering
% \small
% \begin{adjustbox}{angle=90}
\caption{Performance on chinese LLM evaluation with $\mathtt{gpt}$-$\mathtt{4}$ as the neurons.}
\label{tb4}
\begin{tabular}{cccc}
\toprule
&Acc & Macro-F1 & Kap. \\ \midrule
GPT-4 &0.6700 & 0.6261 &0.4587 \\
FairEval & 0.6800 & 0.6692 & 0.5074 \\
WideDeep (Ours) &0.7400 & 0.7245 &0.5965 \\ 
\bottomrule
\end{tabular}
\end{table}

In Table~\ref{tb4}, we present a comparison of the effectiveness of WideDeep, FairEval, and standalone $\mathtt{gpt}$-$\mathtt{4}$ Evaluator in Chinese LLM evaluation. 
WideDeep's advantages have further expanded compared to English benchmarks, with improvements of 6pts, 5.5pts, and 8.9pts in accuracy, F1 score, and kappa correlation coefficient, respectively, achieving a labeling accuracy of 74\%.
The agreement among humans during the Chinese LLM evaluation stands at 80\%, which indicates that WideDeep has reached a 93\% agreement level among humans. 
In fact, with each point increase in accuracy, a significant amount of manual annotation time can be reduced. 
Assuming the LLM evaluator's accuracy is x, the annotators only need to review $\frac{0.8-x}{1-x}$\% of the data annotated by the LLM Evaluator to correct the labeling errors and achieve an 80\% accuracy level, aligning with manual annotation. 
Therefore, the annotators only need to inspect 23\% of the predicted results from WideDeep, while they would have to inspect 37.5\% from FairEval and 39.3\% from GPT-4.
Overall, WideDeep has accelerated the LLM evaluation process by 4.6 times, saving a significant amount of time for human annotators.
Furthermore, the average annotation cost per sample has decreased by 60\%.

\section{Conclusion}
In this paper, we explore whether evaluation performance can be improved in deeper and wider LLM networks. 
Specifically, each neuron within the LLM network assumes a distinct evaluation role, and multiple neurons interact and collaborate, much like the interaction observed in deep neural networks. 
The evaluation process follows a feedforward approach, with each layer of neurons receiving inputs from the previous layer, facilitating a thorough and comprehensive assessment.
An intuitive analogy for our designed LLM network can be drawn to the process of academic paper reviewing. Additionally, we present LLMEval$^2$, the largest and most diverse evaluation benchmark developed to date for the LLM Evaluator. 
Through extensive experiments, we demonstrate that a two-layer wider LLM network yields the best results, significantly enhancing the ability of LLMs to evaluate the quality of generated text.
Furthermore, we apply our evaluator to assess the performance of Chinese LLMs, where it proves to speed up LLM evaluation process by 4.6 times and decrease the average annotation cost per sample by 60\%.

\bibliographystyle{plain}
\bibliography{sample-base}

\begin{thebibliography}{10}

\bibitem{bai2022training}
Yuntao Bai, Andy Jones, Kamal Ndousse, Amanda Askell, Anna Chen, Nova DasSarma,
  Dawn Drain, Stanislav Fort, Deep Ganguli, Tom Henighan, et~al.
\newblock Training a helpful and harmless assistant with reinforcement learning
  from human feedback.
\newblock {\em arXiv preprint arXiv:2204.05862}, 2022.

\bibitem{bau2018identifying}
Anthony Bau, Yonatan Belinkov, Hassan Sajjad, Nadir Durrani, Fahim Dalvi, and
  James Glass.
\newblock Identifying and controlling important neurons in neural machine
  translation.
\newblock {\em arXiv preprint arXiv:1811.01157}, 2018.

\bibitem{bau2020understanding}
David Bau, Jun-Yan Zhu, Hendrik Strobelt, Agata Lapedriza, Bolei Zhou, and
  Antonio Torralba.
\newblock Understanding the role of individual units in a deep neural network.
\newblock {\em Proceedings of the National Academy of Sciences},
  117(48):30071--30078, 2020.

\bibitem{bhaskar2023prompted}
Adithya Bhaskar, Alex Fabbri, and Greg Durrett.
\newblock Prompted opinion summarization with gpt-3.5.
\newblock In {\em Findings of the Association for Computational Linguistics:
  ACL 2023}, pages 9282--9300, 2023.

\bibitem{chang2023survey}
Yupeng Chang, Xu~Wang, Jindong Wang, Yuan Wu, Kaijie Zhu, Hao Chen, Linyi Yang,
  Xiaoyuan Yi, Cunxiang Wang, Yidong Wang, et~al.
\newblock A survey on evaluation of large language models.
\newblock {\em arXiv preprint arXiv:2307.03109}, 2023.

\bibitem{chen2021evaluating}
Mark Chen, Jerry Tworek, Heewoo Jun, Qiming Yuan, Henrique Ponde de~Oliveira
  Pinto, Jared Kaplan, Harri Edwards, Yuri Burda, Nicholas Joseph, Greg
  Brockman, et~al.
\newblock Evaluating large language models trained on code.
\newblock {\em arXiv preprint arXiv:2107.03374}, 2021.

\bibitem{chen2023exploring}
Yi~Chen, Rui Wang, Haiyun Jiang, Shuming Shi, and Ruifeng Xu.
\newblock Exploring the use of large language models for reference-free text
  quality evaluation: A preliminary empirical study, 2023.

\bibitem{fu2023gptscore}
Jinlan Fu, See-Kiong Ng, Zhengbao Jiang, and Pengfei Liu.
\newblock Gptscore: Evaluate as you desire.
\newblock {\em arXiv preprint arXiv:2302.04166}, 2023.

\bibitem{gabeur2020multi}
Valentin Gabeur, Chen Sun, Karteek Alahari, and Cordelia Schmid.
\newblock Multi-modal transformer for video retrieval.
\newblock In {\em Computer Vision--ECCV 2020: 16th European Conference,
  Glasgow, UK, August 23--28, 2020, Proceedings, Part IV 16}, pages 214--229.
  Springer, 2020.

\bibitem{hao2019visualizing}
Yaru Hao, Li~Dong, Furu Wei, and Ke~Xu.
\newblock Visualizing and understanding the effectiveness of bert.
\newblock {\em arXiv preprint arXiv:1908.05620}, 2019.

\bibitem{he2016deep}
Kaiming He, Xiangyu Zhang, Shaoqing Ren, and Jian Sun.
\newblock Deep residual learning for image recognition.
\newblock In {\em Proceedings of the IEEE conference on computer vision and
  pattern recognition}, pages 770--778, 2016.

\bibitem{hendrycks2020measuring}
Dan Hendrycks, Collin Burns, Steven Basart, Andy Zou, Mantas Mazeika, Dawn
  Song, and Jacob Steinhardt.
\newblock Measuring massive multitask language understanding.
\newblock {\em arXiv preprint arXiv:2009.03300}, 2020.

\bibitem{jain2023self}
Siddhartha Jain, Xiaofei Ma, Anoop Deoras, and Bing Xiang.
\newblock Self-consistency for open-ended generations.
\newblock {\em arXiv preprint arXiv:2307.06857}, 2023.

\bibitem{ji2023exploring}
Yunjie Ji, Yan Gong, Yiping Peng, Chao Ni, Peiyan Sun, Dongyu Pan, Baochang Ma,
  and Xiangang Li.
\newblock Exploring chatgpt's ability to rank content: A preliminary study on
  consistency with human preferences, 2023.

\bibitem{karpathy2015visualizing}
Andrej Karpathy, Justin Johnson, and Li~Fei-Fei.
\newblock Visualizing and understanding recurrent networks.
\newblock {\em arXiv preprint arXiv:1506.02078}, 2015.

\bibitem{kocmi2023large}
Tom Kocmi and Christian Federmann.
\newblock Large language models are state-of-the-art evaluators of translation
  quality.
\newblock {\em arXiv preprint arXiv:2302.14520}, 2023.

\bibitem{lee2017going}
Hyungtae Lee and Heesung Kwon.
\newblock Going deeper with contextual cnn for hyperspectral image
  classification.
\newblock {\em IEEE Transactions on Image Processing}, 26(10):4843--4855, 2017.

\bibitem{li2023api}
Minghao Li, Feifan Song, Bowen Yu, Haiyang Yu, Zhoujun Li, Fei Huang, and
  Yongbin Li.
\newblock Api-bank: A benchmark for tool-augmented llms.
\newblock {\em arXiv preprint arXiv:2304.08244}, 2023.

\bibitem{lin-2004-rouge}
Chin-Yew Lin.
\newblock {ROUGE}: A package for automatic evaluation of summaries.
\newblock In {\em Text Summarization Branches Out}, pages 74--81, Barcelona,
  Spain, July 2004. Association for Computational Linguistics.

\bibitem{liu2021hit}
Song Liu, Haoqi Fan, Shengsheng Qian, Yiru Chen, Wenkui Ding, and Zhongyuan
  Wang.
\newblock Hit: Hierarchical transformer with momentum contrast for video-text
  retrieval.
\newblock In {\em Proceedings of the IEEE/CVF International Conference on
  Computer Vision}, pages 11915--11925, 2021.

\bibitem{liu2023gpteval}
Yang Liu, Dan Iter, Yichong Xu, Shuohang Wang, Ruochen Xu, and Chenguang Zhu.
\newblock Gpteval: Nlg evaluation using gpt-4 with better human alignment.
\newblock {\em arXiv preprint arXiv:2303.16634}, 2023.

\bibitem{mahendran2015understanding}
Aravindh Mahendran and Andrea Vedaldi.
\newblock Understanding deep image representations by inverting them.
\newblock In {\em Proceedings of the IEEE conference on computer vision and
  pattern recognition}, pages 5188--5196, 2015.

\bibitem{openai2022chatgpt}
OpenAI.
\newblock Introducing chatgpt.
\newblock 2022.

\bibitem{papineni-etal-2002-bleu}
Kishore Papineni, Salim Roukos, Todd Ward, and Wei-Jing Zhu.
\newblock {B}leu: a method for automatic evaluation of machine translation.
\newblock In {\em Proceedings of the 40th Annual Meeting of the Association for
  Computational Linguistics}, pages 311--318, Philadelphia, Pennsylvania, USA,
  July 2002. Association for Computational Linguistics.

\bibitem{patrick2020support}
Mandela Patrick, Po-Yao Huang, Yuki Asano, Florian Metze, Alexander Hauptmann,
  Joao Henriques, and Andrea Vedaldi.
\newblock Support-set bottlenecks for video-text representation learning.
\newblock {\em arXiv preprint arXiv:2010.02824}, 2020.

\bibitem{qiao2019understanding}
Yifan Qiao, Chenyan Xiong, Zhenghao Liu, and Zhiyuan Liu.
\newblock Understanding the behaviors of bert in ranking.
\newblock {\em arXiv preprint arXiv:1904.07531}, 2019.

\bibitem{radford2017learning}
Alec Radford, Rafal Jozefowicz, and Ilya Sutskever.
\newblock Learning to generate reviews and discovering sentiment.
\newblock {\em arXiv preprint arXiv:1704.01444}, 2017.

\bibitem{saunders2022self}
William Saunders, Catherine Yeh, Jeff Wu, Steven Bills, Long Ouyang, Jonathan
  Ward, and Jan Leike.
\newblock Self-critiquing models for assisting human evaluators.
\newblock {\em arXiv preprint arXiv:2206.05802}, 2022.

\bibitem{song2023preference}
Feifan Song, Bowen Yu, Minghao Li, Haiyang Yu, Fei Huang, Yongbin Li, and
  Houfeng Wang.
\newblock Preference ranking optimization for human alignment.
\newblock {\em arXiv preprint arXiv:2306.17492}, 2023.

\bibitem{sordoni2023deep}
Alessandro Sordoni, Xingdi Yuan, Marc-Alexandre C{\^o}t{\'e}, Matheus Pereira,
  Adam Trischler, Ziang Xiao, Arian Hosseini, Friederike Niedtner, and
  Nicolas~Le Roux.
\newblock Deep language networks: Joint prompt training of stacked llms using
  variational inference.
\newblock {\em arXiv preprint arXiv:2306.12509}, 2023.

\bibitem{touvron2023llama}
Hugo Touvron, Thibaut Lavril, Gautier Izacard, Xavier Martinet, Marie-Anne
  Lachaux, Timothée Lacroix, Baptiste Rozière, Naman Goyal, Eric Hambro,
  Faisal Azhar, Aurelien Rodriguez, Armand Joulin, Edouard Grave, and Guillaume
  Lample.
\newblock Llama: Open and efficient foundation language models, 2023.

\bibitem{wang2023chatgpt}
Jiaan Wang, Yunlong Liang, Fandong Meng, Haoxiang Shi, Zhixu Li, Jinan Xu,
  Jianfeng Qu, and Jie Zhou.
\newblock Is chatgpt a good nlg evaluator? a preliminary study.
\newblock {\em arXiv preprint arXiv:2303.04048}, 2023.

\bibitem{wang2023large}
Peiyi Wang, Lei Li, Liang Chen, Dawei Zhu, Binghuai Lin, Yunbo Cao, Qi~Liu,
  Tianyu Liu, and Zhifang Sui.
\newblock Large language models are not fair evaluators, 2023.

\bibitem{wang2023pandalm}
Yidong Wang, Zhuohao Yu, Zhengran Zeng, Linyi Yang, Cunxiang Wang, Hao Chen,
  Chaoya Jiang, Rui Xie, Jindong Wang, Xing Xie, Wei Ye, Shikun Zhang, and Yue
  Zhang.
\newblock Pandalm: An automatic evaluation benchmark for llm instruction tuning
  optimization, 2023.

\bibitem{wang-etal-2023-self-instruct}
Yizhong Wang, Yeganeh Kordi, Swaroop Mishra, Alisa Liu, Noah~A. Smith, Daniel
  Khashabi, and Hannaneh Hajishirzi.
\newblock Self-instruct: Aligning language models with self-generated
  instructions.
\newblock In {\em Proceedings of the 61st Annual Meeting of the Association for
  Computational Linguistics (Volume 1: Long Papers)}, pages 13484--13508.
  Association for Computational Linguistics, 2023.

\bibitem{wei2023polylm}
Xiangpeng Wei, Haoran Wei, Huan Lin, Tianhao Li, Pei Zhang, Xingzhang Ren, Mei
  Li, Yu~Wan, Zhiwei Cao, Binbin Xie, et~al.
\newblock Polylm: An open source polyglot large language model.
\newblock {\em arXiv preprint arXiv:2307.06018}, 2023.

\bibitem{xie2017aggregated}
Saining Xie, Ross Girshick, Piotr Doll{\'a}r, Zhuowen Tu, and Kaiming He.
\newblock Aggregated residual transformations for deep neural networks.
\newblock In {\em Proceedings of the IEEE conference on computer vision and
  pattern recognition}, pages 1492--1500, 2017.

\bibitem{yuan2023rrhf}
Zheng Yuan, Hongyi Yuan, Chuanqi Tan, Wei Wang, Songfang Huang, and Fei Huang.
\newblock Rrhf: Rank responses to align language models with human feedback
  without tears.
\newblock {\em arXiv preprint arXiv:2304.05302}, 2023.

\bibitem{zeiler2014visualizing}
Matthew~D Zeiler and Rob Fergus.
\newblock Visualizing and understanding convolutional networks.
\newblock In {\em Computer Vision--ECCV 2014: 13th European Conference, Zurich,
  Switzerland, September 6-12, 2014, Proceedings, Part I 13}, pages 818--833.
  Springer, 2014.

\bibitem{zhang2019bertscore}
Tianyi Zhang, Varsha Kishore, Felix Wu, Kilian~Q Weinberger, and Yoav Artzi.
\newblock Bertscore: Evaluating text generation with bert.
\newblock {\em arXiv preprint arXiv:1904.09675}, 2019.

\bibitem{zhao2019moverscore}
Wei Zhao, Maxime Peyrard, Fei Liu, Yang Gao, Christian~M Meyer, and Steffen
  Eger.
\newblock Moverscore: Text generation evaluating with contextualized embeddings
  and earth mover distance.
\newblock {\em arXiv preprint arXiv:1909.02622}, 2019.

\bibitem{zhao2023causal}
Yingxiu Zhao, Bowen Yu, Haiyang Yu, Bowen Li, Chao Wang, Fei Huang, Yongbin Li,
  and Nevin~L Zhang.
\newblock Causal document-grounded dialogue pre-training.
\newblock {\em arXiv preprint arXiv:2305.10927}, 2023.

\bibitem{zheng2023judging}
Lianmin Zheng, Wei-Lin Chiang, Ying Sheng, Siyuan Zhuang, Zhanghao Wu, Yonghao
  Zhuang, Zi~Lin, Zhuohan Li, Dacheng Li, Eric.~P Xing, Hao Zhang, Joseph~E.
  Gonzalez, and Ion Stoica.
\newblock Judging llm-as-a-judge with mt-bench and chatbot arena, 2023.

\bibitem{zhou2014object}
Bolei Zhou, Aditya Khosla, Agata Lapedriza, Aude Oliva, and Antonio Torralba.
\newblock Object detectors emerge in deep scene cnns.
\newblock {\em arXiv preprint arXiv:1412.6856}, 2014.

\end{thebibliography}

% \section*{References}

% References follow the acknowledgments in the camera-ready paper. Use unnumbered first-level heading for
% the references. Any choice of citation style is acceptable as long as you are
% consistent. It is permissible to reduce the font size to \verb+small+ (9 point)
% when listing the references.
% Note that the Reference section does not count towards the page limit.
% \medskip

% {
% \small

% [1] Alexander, J.A.\ \& Mozer, M.C.\ (1995) Template-based algorithms for
% connectionist rule extraction. In G.\ Tesauro, D.S.\ Touretzky and T.K.\ Leen
% (eds.), {\it Advances in Neural Information Processing Systems 7},
% pp.\ 609--616. Cambridge, MA: MIT Press.

% [2] Bower, J.M.\ \& Beeman, D.\ (1995) {\it The Book of GENESIS: Exploring
%   Realistic Neural Models with the GEneral NEural SImulation System.}  New York:
% TELOS/Springer--Verlag.

% [3] Hasselmo, M.E., Schnell, E.\ \& Barkai, E.\ (1995) Dynamics of learning and
% recall at excitatory recurrent synapses and cholinergic modulation in rat
% hippocampal region CA3. {\it Journal of Neuroscience} {\bf 15}(7):5249-5262.
% }

%%%%%%%%%%%%%%%%%%%%%%%%%%%%%%%%%%%%%%%%%%%%%%%%%%%%%%%%%%%%

\end{document}